# AnaMoDiff: 2D Analogical Motion Diffusion via Disentangled Denoising

MAHAM TANVEER, Simon Fraser University, Canada
YIZHI WANG, Simon Fraser University, Canada
RUIQI WANG, Simon Fraser University, Canada
NANXUAN ZHAO, Adobe Research, USA
ALI MAHDAVI-AMIRI, Simon Fraser University, Canada
HAO ZHANG, Simon Fraser University, Canada

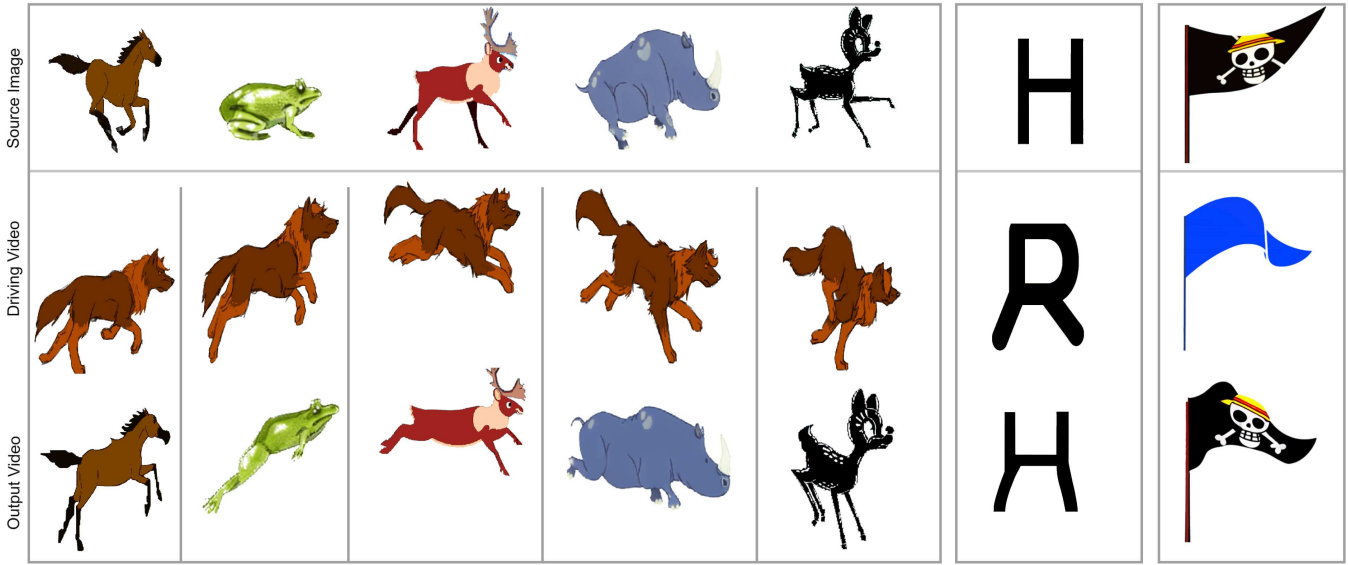

Fig. 1. Our diffusion-based method, AnaMoDiff, performs 2D motion analogy from a driving video (middle row) onto a different object given in a source image (top). Left: a dog motion transferred onto five different animals, displaying one output frame for each animal that matches the corresponding frame from the dog motion. Right: two examples demonstrating our ability to handle non-creature motions.

We present *AnaMoDiff*, a novel *diffusion*-based method for 2D motion analogies that is applied to raw, *unannotated* videos of articulated characters. Our goal is to accurately transfer motions from a 2D driving video onto a source character, with its identity, in terms of appearance and natural movement, well preserved, even when there may be significant discrepancies between the source and driving characters in their part proportions and movement speed and styles. Our diffusion model transfers the input motion via a *latent optical flow* (LOF) network operating in a *noised latent* space, which is spatially aware, efficient to process compared to the original RGB videos, and artifact-resistant through the diffusion denoising process even amid dense movements. To accomplish both motion analogy and identity preservation, we train our denoising model in a *feature-disentangled* manner, operating at two noise levels. While identity-revealing features of the source are learned via conventional noise injection, motion features are learned from LOF-warped videos by only injecting noise with *large values*, with the stipulation that motion properties involving pose and limbs are encoded by higher-level features. Experiments demonstrate that our method achieves the best trade-off between motion analogy and identity preservation.

## 1 INTRODUCTION

Automated synthesis of articulated motions of humans and creatures has been a well-studied problem in computer animation. With the rapid advances in machine learning, data-driven approaches that do not rely on direct, human-in-the-loop motion editing or manipulation have received much attention lately. Neural networks can be trained to learn to generate novel motions of the same character from only a single training sequence [28, 29, 42]. Motion analogy [25, 41] and retargeting [2, 35] are both close variants of the same line of example-based approaches, which can synthesize motions for new characters by mimicking a driving motion sequence involving a similar character.

What is common about most of these methods, especially those designed for 3D motion synthesis and including recent attempts at utilizing powerful diffusion generators to model articulated motions [42, 57], is that they all operate on canonical motion representations, such as joint positions and angles, rather than raw RGB

Authors' addresses: Maham Tanveer, Simon Fraser University, Canada, mta122@sfu.ca; Yizhi Wang, Simon Fraser University, Canada, ywa439@sfu.ca; Ruiqi Wang, Simon Fraser University, Canada, ruiqi_w@sfu.ca; Nanxuan Zhao, Adobe Research, San Jose, USA, nanxuanzhao@gmail.com; Ali Mahdavi-Amiri, Simon Fraser University, Canada, amahdavi@sfu.ca; Hao Zhang, Simon Fraser University, Canada, haoz@sfu.ca.

images or 3D captures. The ability to perform direct motion analysis and synthesis over visual data without relying on fixed templates or explicit motion annotations can greatly improve the versatility and scalability of a method [48, 55, 56, 65].

In this paper, we explore the use of *image diffusion* models for 2D motion analogies operating on *unannotated* videos of articulated motions. Specifically, given a 2D driving video and one or more sample motion frames for a source character $S$, our goal is to produce a new video of $S$ whose motion mimics the driving video while respecting the identity in terms of appearance and natural movement of $S$ as reflected by its motion samples; see Fig. 1. As a key feature of our work, we aim to transfer motions between characters that may exhibit significant discrepancies in their part proportions (e.g., a wolf vs. a giraffe) and movement speed and styles (e.g., a lively horse vs. a slow turtle). However, as naturally expected for motion analogies, we assume some structural consistency between the source and target characters. Other than that, our method is applied directly to raw videos without any motion-related annotations or assumptions on specific structures of the characters.

Our main motivation behind adopting diffusion models [63] for motion synthesis is their superior generative capabilities through denoising. This is highly desirable when the generated motions must reveal object parts unobserved in training data. At the same time, performing scene movements in a *noised* space rather than the original image space has demonstrated success in recent works on image [47] and video editing [9], since the diffusion denoising process can well absorb and rectify the edit distortions and other artifacts introduced into the noisy images. However, the motion transfer task in our work involves more complex movements than existing diffusion-based, sparse scene edits can handle. More critically, as both motion analogy with respect to the driving video and identity preservation with respect to the source character must be fulfilled at the same time, our denoising network must be carefully trained to accommodate both motion-related features and identity-revealing features.

Our proposed solution framework, coined *AnaMoDiff*, for 2D analogical motion diffusion, operates in a *noised latent* space. Specifically, given an initial frame of the source character, we first obtain a corresponding latent code through Stable Diffusion (SD) [43] and then inject noise into the code via DDIM (Denoising Diffusion Implicit Models) [52] inversion. We apply an *optical flow* network, which was trained on SD latent space, to the resulting noised latent code to warp the source character to reproduce motions in the driving video. While prior works [34, 47] have shown that the SD latent space is *spatially aware* to support sparse editing, our work is the first to learn a latent optical flow (LOF) for motion transfer, which involves *dense* movements. Moreover, the latent code provides a reduced image representation by an order of magnitude (from $512^2$ to $64^2$) to improve the efficiency of LOF and diffusion.

Applying conventional denoising, through SD, to the LOF-warped DDIM inverse can produce motions analogous to that of the driving video, but it often fails to preserve the identity of the source character and suffers from other visual artifacts due, in part, to temporal inconsistency. We solve these issues by training our denoising model in a two-level and *feature-disentangled* manner, operating at two noise levels that take on different training inputs:

- First, we fine-tune our denoising network using the source video and conventional (random) noise injection, aiming to learn the appearance and identity-revealing features of the source object.
- Furthermore, we use the LOF-warped video to train the denoising network by only injecting noise with *large values*. We stipulate that motion properties involving pose and limb movements are encoded by higher-level features (e.g., compared to fine-grained textures), which can be better learned at this noise level. Shielding the network training from low-value noise, we can effectively prevent the learned model from introducing lower-level and non-motion related artifacts into the denoised images.

We evaluate our method both qualitatively and through a user study, comparing with state-of-the-art alternatives. The results show that AnaMoDiff offers the best trade-off between motion transfer and identity preservation, with minimal visual artifacts. We also demonstrate the applicability of our method beyond articulated humans and creatures (e.g., letter shapes) and motions (e.g., flag waving).

## 2 RELATED WORK

Here, we first discuss general motion retargeting and transfer, as we are transferring motion from one video to another. Our method is based on a pre-trained image diffusion model. Therefore, we discuss techniques for video and animation that benefit from diffusion models. Lastly, since an important part of our method is to predict an optical flow on the latent space, we also discuss image-to-video-generation techniques that utilize an optical flow in their process.

**Motion Retargeting/Transfer.** Motion retargeting aims to transfer movements from a source character to a target character while adapting them to the specific characteristics of the target. The problem was initially proposed by Gleicher [13] and traditional methods [20, 25, 36, 53] involve retargeting between joints (skeletons) with the use of inverse kinematic solvers. Currently, deep learning techniques are mostly used to extract features from skeletons and then learn to adapt them to the given character [1, 3, 8, 21]. Most of the above-mentioned methods target 3D body motions where the datasets inherently contain skeleton information. However, for general 2D videos and motions, acquiring object skeletons necessitates annotations, and the process is challenging since 2D characters and their skeletons are not pre-defined, and other challenges, such as occlusion, exist. This limits the application of these methods on general 2D video datasets. To remedy this, RecycleGAN [5] accomplishes unsupervised 2D motion retargeting by extending image-to-image translation frameworks (e.g., Cycle-GAN [67]) to handle videos without relying on skeleton supervision. However, it struggles to align objects with substantially different shapes. Increasingly, methods aim to extract key-points from objects in a self-supervised or unsupervised manner instead of pre-defined skeletons [17, 18]. JOKR [35] introduces cross-domain motion retargeting, i.e., the source and target videos are of different shapes, using a joint key-point representation. TPS-Motion-Model [65] proposes thin-plate spline motion estimation to produce an optical flow warping the feature maps of the source to those of the driving image.

These methods are trained on datasets featuring specific object motions. To enable generalization across arbitrary motions, our method, AnaMoDiff generates the motion frames based on SD to leverage its strong prior knowledge. In contrast to calculating optical flows in pixel space, AnaMoDiff performs this computation in the SD latent space. This approach allows for a more compact representation of both object appearance and motions, facilitating effective motion transfer.

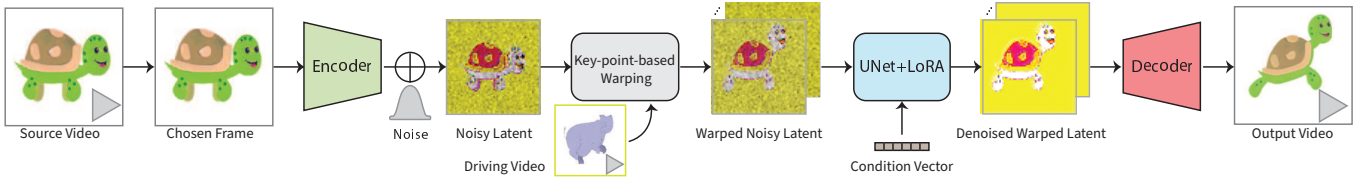

Fig. 2. Given a source video, a frame (e.g., the first one), is selected and mapped to the noisy latent space. A key-point-based warping is applied to the noisy latent to transfer the motion of the driving video to the chosen frame. The resulting warped noisy latents along with a condition vector obtained from a text prompt (e.g., turtle) are sent to a UNet fine-tuned via LoRA. The resulting denoised warped latents are sent to the SD's decoder to produce the output video.

**Diffusion for Video and Animation Synthesis.**

After the great success of text-to-image diffusion models, recently, video diffusion models have been introduced to generate a video from text prompts [4, 6, 7, 10, 14, 19, 22, 30, 32, 51, 59, 66]. These models are capable of producing realistic and high-quality but usually short videos. However, they are not readily suitable for motion transfer. We instead use pre-trained Stable Diffusion [43], fine-tuned to replicate the natural movements of the source video. Our method shares similarities with personalization techniques for diffusion models [11], including those involving fine-tuning [15, 27, 37, 45]. Similar to our method, fine-tuning methods exist that use LoRA for personalization [15, 24, 46]. Tune-A-Video [61] extends image personalization to videos. Our fine-tuning method, however, is unique. We fine-tune Stable Diffusion with two inputs, original and warped motions of objects, ensuring generated results preserve both object identity and an extended range of motions.

In addition to general text-to-video generation models, some techniques have been introduced for more targeted applications. Animatediff [16] performs personalized text-to-video synthesis by distilling generalizable motion priors from large video datasets. LAMP [62] learns a motion pattern from a small video set and generates videos with the learned motion. Gal et al. [12] animates a still sketch in vector format according to a text prompt describing a desired action.

Due to the importance of human movements and activities, many techniques have also been proposed for human animation. Tevet et al. [57] introduces an adapted classifier-free diffusion-based generative model for 3D human motion synthesis. LEO [60] utilizes a Latent Diffusion Model (LDM) [43] to generate a sequence of flow maps, which are used to warp and inpaint the starting frame to produce the video sequence. For pose-guided image-to-video synthesis for character animation, Hu et al. [24] proposed an SD-based method that considers the consistency of intricate appearance features from human images and the controllability and temporal continuity of synthesized poses. Dreampose [26] adapts SD Model to enable image and pose conditioning. Disco [58] enhances the generalizability of dance motion synthesis via disentangled control of pose and background.

Contrary to approaches that involve a fixed skeleton, our method employs a technique free from such constraints. Our method applies optical flow within the latent space of a pre-trained stable diffusion and incorporates a disentangled denoising strategy. As a result, our approach effectively maintains the driving motion while also preserving the motion and appearance details of the source video.

**Image-to-Video Generation via Optical Flow.** Optical flows have been employed to guide generative models in the task of transforming still images into dynamic videos [31, 33, 39, 64].

LFDM [38] generates videos by warping images using flow sequences generated in the latent space, conditioned on class labels. It generates optical flows for specific actions using a dataset with various subjects performing specific actions like crossing arms. The model is limited to specific learned motions, such as subjects exhibiting full-standing human bodies or centered and focused faces.

In contrast to LFDM, our acquired optical flows can accommodate diverse motions. Additionally, we introduce a disentangled denoising approach (Section 3.3) to fine-tune SD [44], enhancing the output motion by preserving the appearance and natural movements of the source object.

Recently, DragGAN [40] and DragDiffusion [47] drag a point on an image to another location to make edits, which can produce a motion of the select region. However, formulating an intricate motion in this setting, especially for our application of motion transfer, is challenging and not straightforward. We instead learn optical flow through some key-points to better define the object movements and produce coherent motions.

## 3 METHOD

Here, we introduce our method, AnaMoDiff, which transfers motion from a driving video to a source video without requiring annotation. This process aims to maintain the identity of the object in the source video, ensuring consistency in both appearance and natural movement. AnaMoDiff operates in the noisy latent space of Stable Diffusion (SD). Relying on the generative capabilities of a pre-trained SD enables propagating the motion properties while avoiding undesired artifacts and deformations. AnaMoDiff, first, applies a key-point-based warping to the noisy latent code of a chosen frame (e.g., first frame) of a source video. This warping is to modify the frame according to the motion of a driving video (Fig. 2). After warping, the resulting warped noisy latents are denoised using the SD's denoiser (i.e., UNet) that has been fine-tuned via LoRA [23]. Finally, denoised warped latents are mapped back to the spatial domain using SD's decoder to produce the output video.

Warping includes pre-training our Latent Optical Flow Network (LOFNet) that involves mapping a random frame to other frames within a single video (See Fig. 3 and Section 3.2). During inference, a frame from a source video is warped in noise space according to the motion of a driving video whose frames' key points are adapted and aligned to the key points of the source object (see Fig. 4).

Furthermore, to remain faithful to the characteristics and movements of the source video, the UNet is fine-tuned using LoRA based on both the source video and an intermediary warped version of the source object (see Section 3.3). We train the UNet in two phases, using distinct noise levels and inputs (see Fig. 4). Initially, we fine-tune it with source video using random noise injection to capture the source object's appearance. Then, we employ the warped noisy latents obtained from LOFNet, utilizing noise with high values to only focus on plausible motion properties.

Formally, AnaMoDiff takes as input a source video $V_S$ and a driving video $V_D$. The object in $V_S$ to be animated is referred to as $O_S$, which takes its motion from object $O_D$ in $V_D$ to produce a plausible and realistic output video $W_s$. We use the SD's encoder to convert the frames of videos $F_S^i$ and $F_D^i$ into latent space $Z_S^i \in Z_S$ and $Z_D^i \in Z_D$, where $Z_S$ and $Z_D$ represent the latent space of $V_S$ and $V_D$. Our method then uses a warping network, LOFNet, to find an optical flow motion for $Z_D$. This is done using corresponding key-points. Afterwards, we adapt it to the key-points of $Z_S$. Directly applying this to $F_S^0$'s latent ($Z_S^0$), we get an intermediate representation of our output video, which we call $Z_{WI}$ in latent, or its image counter-part as $W_I$. We then fine-tune SD to maintain the recognizability of $O_S$, for which we use a disentangled two-level training method discussed in Section 3.3. This can generate $O_S$ in the required motion with identity preservation, and temporal coherency. In the following, we first provide a review of diffusion models and then discuss our method's components.

## 3.1 Preliminaries on Stable Diffusion

**Stable Diffusion.** Latent diffusion models (LDM) [43] are a class of diffusion models trained to denoise a latent representation as opposed to images directly. This greatly reduces the computational needs of the diffusion models and has made models like Stable Diffusion (SD) based on LDM very popular. In LDMs, an autoencoder is trained to encode high-resolution images into a compact latent representation. We use SD model v1.4, which has a latent resolution of 64x64 with 4 channels, encoding RGB images of resolution 512x512.

**Stable Diffusion for Video Generation.** SD employs a UNet for the denoising process, which is composed of 2D residual blocks and transformer blocks. The transformer blocks consist of a spatial layer followed by a cross-attention layer (for the conditional inputs; e.g. text). Since it is designed for generating images, some modifications are necessary to adapt it for video generation. To do so, we employ Wu et al. [61]'s architecture for inflating the 2D convolution layers to pseudo-3D convolution layers, replacing the $3 \times 3$ kernels with $1 \times 3 \times 3$, appending a temporal transformer layer in each transformer block and enhancing the spatial transformer to spatio-temporal. For architecture details, consult the original paper.

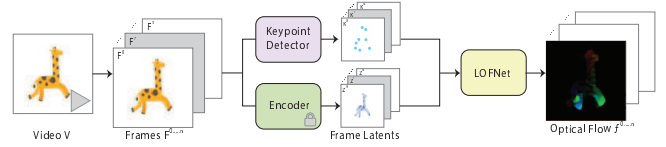

Fig. 3. Step 1: LofNet is trained on latent input of our data of articulated animals to generate an optical flow, $f$. It receives a set of key-points and latent frames and learns to map two random frames.

**LoRA in Diffusion Models.** LoRA [23] efficiently fine-tunes large networks by freezing network weights and introducing trainable rank decomposition networks into each layer. The assumption is that the model weight adaptation happens within a low-rank subspace. Notably, it has gained attention in diffusion models for its efficiency [46, 47]. In our method, 150-200 iterations usually suffice for plausible results; however, to cover complex cases, we employ a consistent 250 iterations across all examples.

## 3.2 Step 1: Training LOFNet

We train a warper network, Latent Optical Flow Network (LOFNet), to estimate the flow $f$ between the latents of two frames (Fig. 3). In each iteration, a reference frame $F_r$ and a driving frame $F_d$, are randomly chosen from the same video. A pre-trained key-point detector, initially loaded with weights from [48] estimates key-points $K_i$ and their jacobians on RGB frames that are passed to LOFNet. A set of sparse trajectories is generated using the pre-computed key-points. Motion is modeled in the vicinity of each key-point with key-point displacements and local affine transformations. An encoder-decoder similar to [49, 50, 65] is then used to generate optical flow $f$ using these local motion approximations. $f$ has two channels for movements in $x$ and $y$ directions. Applying $f$ to latent $z_r$ of $F_r$ outputs $z_o$, which is compared with the latent of $F_d$, $z_d$, via MSE reconstruction loss: $\mathcal{L}_{warp} = \mathcal{L}_{rec}(z_d, z_o)$. MGIF [49], including $\approx 1,000$ videos of animals, is used to train LOFNet.

Using optical flow in the latent space reduces computation since it is a compact version of the high-resolution image space. It is important to confirm if spatial changes in latent space map effectively to RGB space. Methods such as [34, 47] show that SD's latent space aligns well with its RGB representation, making it suitable for spatial editing. Applying this to complex "editing," we find SD's latent space suitable for dense optical flow, confirming effective motion translation between objects.

## 3.3 Step 2: Disentangled Denoising

Step 2 involves fine-tuning the UNet via LoRA [23] to properly learn appearance and motion features. The denoising network receives two noisy inputs along with a condition vector capturing a text prompt (Fig.4). The prompt describing $O_S$ by few words, such as "black cat," "grey horse," or "giraffe," serves the dual purpose of accelerating the fine-tuning process for LoRA and capitalizing on the learned priors of SD representing specific concepts.

Our primary objective is to extract low-level features of ($O_S$) from the latent representation of $V_S$, ($Z_S$), ensuring the object remains recognizable in the output video. $Z_S$ contributes to learning temporal coherency through the SD's UNet. However, relying solely on $Z_S$ leads to a limited model struggling to generate samples beyond poses and motions in the source video. To address this, we incorporate learning the "expected poses" of the final motion while emphasizing $Z_S$ for optimal low-level feature extraction from $O_S$. Our intuition is guided by the observation that the noise level in diffusion significantly influences the extracted features. Inspired by methods such as [54], we propose restricting the noise level in input to high values to confine diffusion models to extracting high-level features.

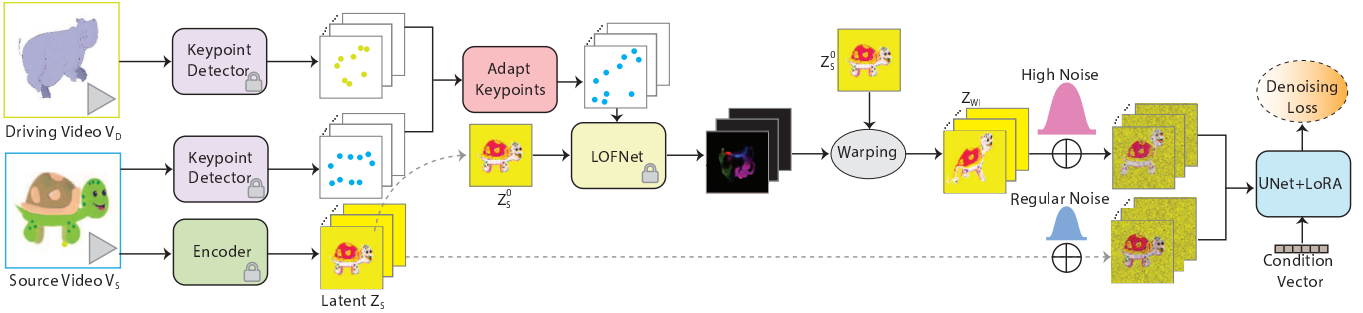

Fig. 4. Using the optical flow generated by LOFNet, we fine-tune the UNet with LoRA, using two inputs. The first input is the latents of our source video, $Z_S$. The second input, $Z_{WI}$, is the first frame of latent $Z_S$, $Z_S^0$ warped to motion of $V_D$. To get the required optical flow motion, we first find a set of key-points that follow motion of $V_D$ but have the shape of $V_S$. For this, we adapt the key-points from $V_D$ to shape of $V_S$ using ConvexHulls to get a rough estimation of the shapes. The adapted key-points along with $Z_S^0$ are then used to generate optical flow via LOFNet, the warped output of which is $Z_{WI}$. For fine-tuning UNet $Z_{WI}$ is trained at only high noise, enforcing UNet to learn low-level features from $Z_S$. The condition vector is the output of CLIP text encoder, with a simple prompt describing $O_S$ (e.g., "a turtle").

To achieve this, we employ an intermediate warped output, denoted as $Z_{WI}$, obtained by warping the latent ($z_r$) of the first frame of $Z_S$ in latent space, guided by the motion of $Z_D$. While $Z_{WI}$ aligns with the motion of $V_D$, it may exhibit artifacts, as illustrated in Fig. 7. While using it directly as the final output is impractical, we leverage its information to learn the features of the "new poses" of $O_S$.

Our proposed method introduces a "disentangled denoising" technique, fine-tuning the diffusion model on two inputs with different noise ranges (see Fig.4). The primary input is $Z_S$, crucial for learning spatial and temporal features of $O_S$. $Z_S$ undergoes training in a regular denoising manner, with added noise ranging from 0 to 1,000. The secondary input is the intermediate warped output, $Z_{WI}$, introduced to the diffusion model only at high noise steps. This combined training approach enables the diffusion model to capture the low-level features of $O_S$ while integrating high-level features from $Z_{WI}$. With that in mind, we can define the loss function as follows. Let diffusion denoising loss be defined as $D = E_{z,\epsilon \sim \mathcal{N},t,c}[||\epsilon - \epsilon_\theta(z,t,c)||_2^2]$ where $\epsilon$ is input noise, $c$ is the CLIP embedding of the input prompt, $\epsilon_\theta$ is the predicted noise and $t$ is the timestep. Let $D_s$ be the denoising loss for $Z_S$ input, and $D_w$ be the denoising loss for $Z_{WI}$. The total loss is then defined as $D_{loss} = D_s + D_w$.

### 3.4 Step 3: DDIM Inverse Warping

LOFNet and UNet are trained and fine-tuned, then the final output is generated. Since source videos are not directly generated by a diffusion model, an inversion step is needed to find the right noise to work with the given videos. A DDIM inverse is applied to $F_S$ to convert the image into noisy latent space, on which $f$ is then applied. Finally, we use the fine-tuned UNet to denoise the warped DDIM inverse to get the final video $W_S$. Figures 5 and 6 demonstrate frames generated by our method in comparison with baselines. More results are provided at the end.

### 3.5 Training Details

We employ an Adam optimizer with a learning rate of $5e^{-4}$ for all components throughout training. Stable Diffusion and Keypoint generator hyperparameters [48] remain at default values. LOFNet is trained for 10,000 epochs with locally saved latent outputs, reducing training time to approximately 25 mins. The noise step range for $Z_W^I$ is set between $850 - 1,000$, with additional noise applied to the warped DDIM inverse for diversity. Histogram matching corrects off-tone outputs, and the UNet is fine-tuned for each $V_D$ and $V_S$ combination. Each experiment involves training and generating 10 frames, taking around seven minutes per experiment on an NVIDIA GeForce RTX 3090.

## 4 RESULTS AND COMPARISONS

To evaluate AnaMoDiff 's motion transfer capabilities, we conduct both qualitative and quantitative assessments. Qualitatively, we visually compare various shape and motion combinations against three baselines. Quantitatively, we conduct two user studies to assess motion transfer quality and the preservation of the source object's appearance. Additionally, we perform a CLIP-based test to ensure frame consistency across textual and image domains.

**Baselines.** We compare our method with three baselines: (1) Thin-Plate Spline Motion Model for Image Animation (TPS) [65], an optical flow-based motion transfer for RGB videos. (2) JOKR [35], a method employing joint keypoint representation for motion retargeting. (3) TuneAVideo [61], a diffusion-based video generation method that uses a text prompt for output specification. Evaluation spans qualitative and quantitative aspects, focusing on (1) Quality of motion transfer and (2) Retaining source object recognizability. TPS is used in "relative" mode to maintain physical recognizability, aligning with our method's relative motion approach. TuneAVideo is trained on our $V_D$, with the text prompt representing the source video (e.g., "a rhino is running"). For inference with a different object, the prompt is adjusted accordingly (e.g., "a green turtle is running").

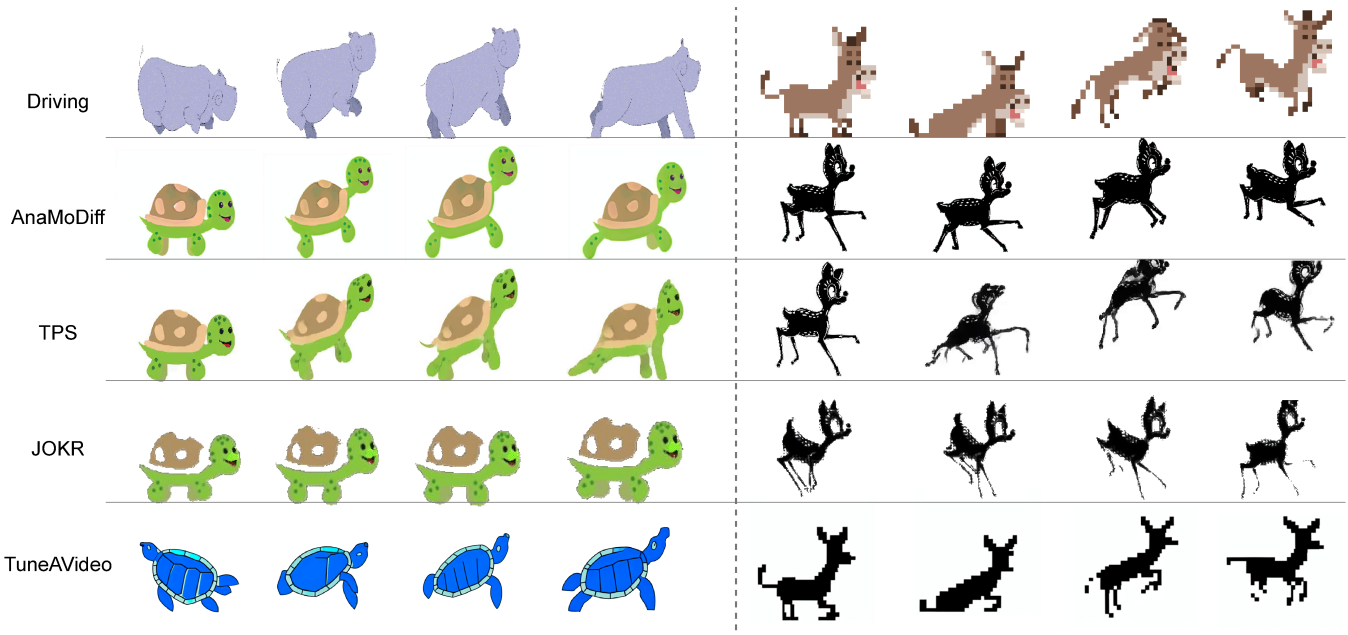

Fig. 5. Animal results compared with: TPS, JOKR, TuneAVideo, where TuneAVideo utilizes prompts like the animal name and color (e.g., "green turtle") along with the motion (e.g., "is running"). Our approach excels in preserving both the character's motion and appearance, resulting in fewer undesirable deformations.

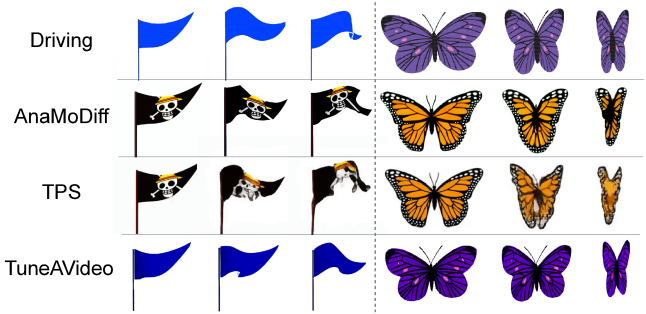

Fig. 6. In-the-wild results. For TuneAVideo's inference, we use prompts such as "a pirate flag is flying". Our approach remains more faithful to the source image in terms of appearance while it also follows the motion provided by the driving video.

| Method | Shape Survey ↓ | Motion Survey ↓ | CLIP-Image ↑ | CLIP-Text ↑ |
| --- | --- | --- | --- | --- |
| AnaMoDiff | **1.39** | **1.53** | 9.397 | **3.147** |
| TPS | 2.02 | 1.83 | 9.141 | 3.134 |
| JOKR | 2.57 | 2.62 | – | – |
| TuneAVideo | – | – | **9.454** | 3.089 |

Table 1. Quantitative results of surveys and CLIP scores.

### 4.1 Qualitative Evaluation

**Animal/Creature Results.**

In Fig. 5, qualitative results on animal inputs are presented. We observe that JOKR struggles when the target motion significantly differs from the original motion domain of $V_S$, as seen in the Turtle-Rhino example where the turtle remains unchanged due to the absence of a jump motion in $V_S$. TPS faces difficulties in preserving the structure of $O_S$ due to its lack of understanding of the object's form. Additionally, diffusion-based methods relying solely on the text input encounter challenges in defining the desired output accurately and often generate results based on their inherent understanding or their default representations of the objects involved. For instance, in the Turtle-Rhino case, despite the prompt being "A green turtle is running," the method persists in generating a "blue" turtle.

**Letter Results.** To demonstrate the versatility and generalizability of our method, we provide some results of transferring motion from animals to text and text to text in Fig. 10. For text, we find that higher noise works better, so we increase our noise steps to 900-1,000. Our approach, unlike TPS, does not produce unnatural artifacts such as the legs at the bottom of the letters. Therefore, our method can transfer motion retaining the characteristics of $O_S$. TuneAVideo also ignores the letters' geometry and appearance. JOKR is not included as it requires the source to be a motion.

**Custom Results.** To further test the generality of our method, we test it on in-the-wild examples. Similar to letters, here, the target is only an image. Fig. 6 shows results on a flag and a butterfly.

### 4.2 Quantitative Evaluation

We perform a quantitative evaluation on MGIF. In the first survey, participants rank motion transfers (AnaMoDiff, JOKR, TPS) based on mimicking driving video motion while maintaining source image quality. The second survey focuses on shape recreation, with participants ranking videos based on preserving source image appearance and minimizing artifacts or noise. Using 15 samples (7 from TPS and JOKR), our results in Table 1, from surveys with ≈ 100 participants each, indicate our approach excels in shape preservation and surpasses baselines when combined with motion.

Fig. 7. Different noise levels for tuning with intermediate results affect the output. "Without $Z_{WI}$", refers to not using intermediate training at all, and just using source video to train. As it is apparent, 800-1,000 produces the best outcome with less noise and more natural poses.

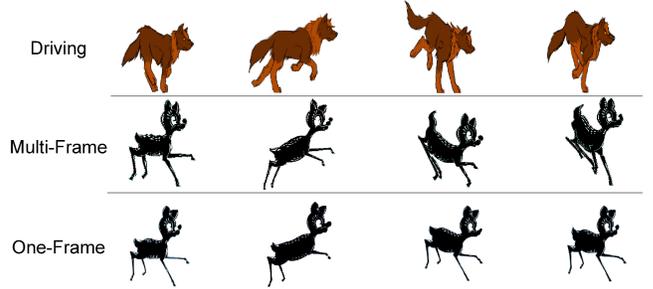

Fig. 8. Single frame vs a video as input. While the method can coarsely follow the motion when the input is a single frame, having a video as input greatly improves the dynamics of the output.

To assess video consistency, we utilize a clip-based method similar to [61]. Using CLIP image embeddings on the first frame, we calculate cosine similarity with every subsequent frame to evaluate generation consistency. Additionally, we employ a text-based CLIP score using the prompt template "2D Illustration of a $O$," replacing $O$ with the target subject (e.g., turtle, cat). This ensures the recognition of the target object in each frame. We conduct tests for five diverse motions (jump, walk, run, etc.), generating 10 examples for each motion with varied source videos. Each video comprises 10 frames, and results are presented in Table 1. We exclude JOKR from comparison due to lengthy experiment times and logistical challenges with a large set (50).

In CLIP-Text analysis, our method excels in maintaining the recognizability of $O_S$, particularly compared to TuneAVideo, which may struggle with generating a target subject distinct from the driving video. Even though CLIP-Image indicates TuneAVideo's better performance, we note that the test assesses consistency within a video rather than concerning $O_S$. The fact that our method closely aligns with a pure diffusion method without restraints highlights its effectiveness in sustaining quality in consistent video generation.

### 4.3 Ablation Studies

**Noise effect on two input training.** Fig. 7 demonstrates the effect of using different noise levels for intermediate warped video in fine-tuning UNet. At lower noise levels, the model starts learning the distortions in low-level features, which greatly affects the outputs. This is highly prominent in the bottom row, where the legs show undesirable blurriness. Without using $Z_{WI}$ in training, however, the model struggles with generating samples outside the range of $V_S$ such as the top row where a jump motion cannot be produced as there is no such motion in the source video $V_S$. The bottom row can work without intermediate training, but only because $V_S$ for the giraffe video, already has a pose similar to the target.

**Effect of using one-frame as input vs video.** Fig. 8 shows the effect of using a video vs a single frame as source input. Although the method can work with a single frame, the output motion can be restricted depending on the source object's complexity. Therefore, if available, an input source video is beneficial to produce a more dynamic output video.

## 5 CONCLUSIONS

We introduced AnaMoDiff to address the task of synthesizing articulated motions from unannotated videos while preserving the identity of the source character. We utilized Stable Diffusion (SD), coupled with a latent optical flow (LOF), to achieve motion transfer between characters with notable structural discrepancies. Our approach can transfer the motions while preserving the identity by fine-tuning SD in a feature-disentangled manner.

The versatility of our method is demonstrated across a range of scenarios, including articulated creatures, humans, letters, and inanimate objects like flags. Through qualitative evaluations and user studies, AnaMoDiff exhibits a compelling trade-off between motion transfer and identity preservation, outperforming SOTA alternatives with minimal visual artifacts. There are still limitations and future directions that need to be explored. Currently, some similarities in starting pose are required for a reasonable outcome. Future work can include a more general motion transfer free of such restraints. Our method may also struggle with single image input as the source. A method to generalize based on a single input is also a possible future work.

# REFERENCES

[1] Kfir Aberman, Peizhuo Li, Dani Lischinski, Olga Sorkine-Hornung, Daniel Cohen-Or, and Baoquan Chen. 2020. Skeleton-aware networks for deep motion retargeting. *ACM Transactions on Graphics (TOG)* 39, 4 (2020), 62–1.

[2] Kfir Aberman, Rundi Wu, Dani Lischinski, Baoquan Chen, and Daniel Cohen-Or. 2019. Learning Character-Agnostic Motion for Motion Retargeting in 2D. *ACM Trans. Graph. (SIGGRAPH)* (2019).

[3] Kfir Aberman, Rundi Wu, Dani Lischinski, Baoquan Chen, and Daniel Cohen-Or. 2019. Learning character-agnostic motion for motion retargeting in 2d. *arXiv preprint arXiv:1905.01680* (2019).

[4] Jie An, Songyang Zhang, Harry Yang, Sonal Gupta, Jia-Bin Huang, Jiebo Luo, and Xi Yin. 2023. Latent-shift: Latent diffusion with temporal shift for efficient text-to-video generation. *arXiv preprint arXiv:2304.08477* (2023).

[5] Aayush Bansal, Shugao Ma, Deva Ramanan, and Yaser Sheikh. 2018. Recycle-gan: Unsupervised video retargeting. In *Proceedings of the European conference on computer vision (ECCV)*. 119–135.

[6] Andreas Blattmann, Tim Dockhorn, Sumith Kulal, Daniel Mendelevitch, Maciej Kilian, Dominik Lorenz, Yam Levi, Zion English, Vikram Voleti, Adam Letts, et al. 2023. Stable video diffusion: Scaling latent video diffusion models to large datasets. *arXiv preprint arXiv:2311.15127* (2023).

[7] Andreas Blattmann, Robin Rombach, Huan Ling, Tim Dockhorn, Seung Wook Kim, Sanja Fidler, and Karsten Kreis. 2023. Align your latents: High-resolution video synthesis with latent diffusion models. In *Proceedings of the IEEE/CVF Conference on Computer Vision and Pattern Recognition*. 22563–22575.

[8] Caroline Chan, Shiry Ginosar, Tinghui Zhou, and Alexei A Efros. 2019. Everybody dance now. In *Proceedings of the IEEE/CVF international conference on computer vision*. 5933–5942.

[9] Yufan Deng, Ruida Wang, Yuhao Zhang, Yu-Wing Tai, and Chi-Keung Tang. 2023. DragVideo: Interactive Drag-style Video Editing. arXiv:2312.02216 [cs.GR]

[10] Patrick Esser, Johnathan Chiu, Parmida Atighehchian, Jonathan Granskog, and Anastasis Germanidis. 2023. Structure and content-guided video synthesis with diffusion models. In *Proceedings of the IEEE/CVF International Conference on Computer Vision*. 7346–7356.

[11] Rinon Gal, Yuval Alaluf, Yuval Atzmon, Or Patashnik, Amit H Bermano, Gal Chechik, and Daniel Cohen-Or. 2022. An image is worth one word: Personalizing text-to-image generation using textual inversion. *arXiv preprint arXiv:2208.01618* (2022).

[12] Rinon Gal, Yael Vinker, Yuval Alaluf, Amit H Bermano, Daniel Cohen-Or, Ariel Shamir, and Gal Chechik. 2023. Breathing Life Into Sketches Using Text-to-Video Priors. *arXiv preprint arXiv:2311.13608* (2023).

[13] Michael Gleicher. 1998. Retargetting Motion to New Characters. In *Proceedings of the 25th Annual Conference on Computer Graphics and Interactive Techniques (SIGGRAPH '98)*. Association for Computing Machinery, New York, NY, USA, 33–42. https://doi.org/10.1145/280814.280820

[14] Jiaxi Gu, Shicong Wang, Haoyu Zhao, Tianyi Lu, Xing Zhang, Zuxuan Wu, Songcen Xu, Wei Zhang, Yu-Gang Jiang, and Hang Xu. 2023. Reuse and diffuse: Iterative denoising for text-to-video generation. *arXiv preprint arXiv:2309.03549* (2023).

[15] Yuchao Gu, Xintao Wang, Jay Zhangjie Wu, Yujun Shi, Yunpeng Chen, Zihan Fan, Wuyou Xiao, Rui Zhao, Shuning Chang, Weijia Wu, et al. 2023. Mix-of-Show: Decentralized Low-Rank Adaptation for Multi-Concept Customization of Diffusion Models. *arXiv preprint arXiv:2305.18292* (2023).

[16] Yuwei Guo, Ceyuan Yang, Anyi Rao, Yaohui Wang, Yu Qiao, Dahua Lin, and Bo Dai. 2023. Animatediff: Animate your personalized text-to-image diffusion models without specific tuning. *arXiv preprint arXiv:2307.04725* (2023).

[17] Xingzhe He, Bastian Wandt, and Helge Rhodin. 2021. Latentkeypointgan: Controlling gans via latent keypoints. *arXiv preprint arXiv:2103.15812* (2021).

[18] Xingzhe He, Bastian Wandt, and Helge Rhodin. 2022. Autolink: Self-supervised learning of human skeletons and object outlines by linking keypoints. *Advances in Neural Information Processing Systems* 35 (2022), 36123–36141.

[19] Yingqing He, Tianyu Yang, Yong Zhang, Ying Shan, and Qifeng Chen. 2022. Latent video diffusion models for high-fidelity video generation with arbitrary lengths. *arXiv preprint arXiv:2211.13221* (2022).

[20] Chris Hecker, Bernd Raabe, Ryan W Enslow, John DeWeese, Jordan Maynard, and Kees van Prooijen. 2008. Real-time motion retargeting to highly varied user-created morphologies. *ACM Transactions on Graphics (TOG)* 27, 3 (2008), 1–11.

[21] Tobias Hinz, Matthew Fisher, Oliver Wang, Eli Shechtman, and Stefan Wermter. 2022. Charactergan: Few-shot keypoint character animation and reposing. In *Proceedings of the IEEE/CVF Winter Conference on Applications of Computer Vision*. 1988–1997.

[22] Jonathan Ho, William Chan, Chitwan Saharia, Jay Whang, Ruiqi Gao, Alexey Gritsenko, Diederik P Kingma, Ben Poole, Mohammad Norouzi, David J Fleet, et al. 2022. Imagen video: High definition video generation with diffusion models. *arXiv preprint arXiv:2210.02303* (2022).

[23] Edward J Hu, Yelong Shen, Phillip Wallis, Zeyuan Allen-Zhu, Yuanzhi Li, Shean Wang, Lu Wang, and Weizhu Chen. 2021. Lora: Low-rank adaptation of large language models. *arXiv preprint arXiv:2106.09685* (2021).

[24] Li Hu, Xin Gao, Peng Zhang, Ke Sun, Bang Zhang, and Liefeng Bo. 2023. Animate Anyone: Consistent and Controllable Image-to-Video Synthesis for Character Animation. *arXiv preprint arXiv:2311.17117* (2023).

[25] Sumit Jain and C. Karen Liu. 2009. Motion Analogies: Automatic Motion Transfer to Different Morphologies. In *Proc. of Symp. of Computer Animation (SCA)*.

[26] Johanna Karras, Aleksander Holynski, Ting-Chun Wang, and Ira Kemelmacher-Shlizerman. 2023. Dreampose: Fashion image-to-video synthesis via stable diffusion. *arXiv preprint arXiv:2304.06025* (2023).

[27] Nupur Kumari, Bingliang Zhang, Richard Zhang, Eli Shechtman, and Jun-Yan Zhu. 2023. Multi-concept customization of text-to-image diffusion. In *Proceedings of the IEEE/CVF Conference on Computer Vision and Pattern Recognition*. 1931–1941.

[28] Peizhuo Li, Kfir Aberman, Zihan Zhang, Rana Hanocka, and Olga Sorkine-Hornung. 2022. GANimator: Neural Motion Synthesis from a Single Sequence. *ACM Trans. Graph. (SIGGRAPH)* (2022).

[29] Weiyu Li, Xuelin Chen, Peizhuo Li, Olga Sorkine-Hornung, and Baoquan Chen. 2023. Example-based Motion Synthesis via Generative Motion Matching. *ACM Trans. Graph. (SIGGRAPH)* (2023).

[30] Xin Li, Wenqing Chu, Ye Wu, Weihang Yuan, Fanglong Liu, Qi Zhang, Fu Li, Haocheng Feng, Errui Ding, and Jingdong Wang. 2023. Videogen: A reference-guided latent diffusion approach for high definition text-to-video generation. *arXiv preprint arXiv:2309.00398* (2023).

[31] Yijun Li, Chen Fang, Jimei Yang, Zhaowen Wang, Xin Lu, and Ming-Hsuan Yang. 2018. Flow-grounded spatial-temporal video prediction from still images. In *Proceedings of the European Conference on Computer Vision (ECCV)*. 600–615.

[32] Zhengxiong Luo, Dayou Chen, Yingya Zhang, Yan Huang, Liang Wang, Yujun Shen, Deli Zhao, Jingren Zhou, and Tieniu Tan. 2023. VideoFusion: Decomposed Diffusion Models for High-Quality Video Generation. In *Proceedings of the IEEE/CVF Conference on Computer Vision and Pattern Recognition*. 10209–10218.

[33] Aniruddha Mahapatra and Kuldeep Kulkarni. 2022. Controllable animation of fluid elements in still images. In *Proceedings of the IEEE/CVF Conference on Computer Vision and Pattern Recognition*. 3667–3676.

[34] Jiafeng Mao, Xueting Wang, and Kiyoharu Aizawa. 2023. Guided Image Synthesis via Initial Image Editing in Diffusion Model. *ACM Int. Conf. Multimedia*.

[35] Ron Mokady, Rotem Tzaban, Sagie Benaim, Amit H Bermano, and Daniel Cohen-Or. 2021. Jokr: Joint keypoint representation for unsupervised cross-domain motion retargeting. *arXiv preprint arXiv:2106.09679* (2021).

[36] Jean-Sébastien Monzani, Paolo Baerlocher, Ronan Boulic, and Daniel Thalmann. 2000. Using an intermediate skeleton and inverse kinematics for motion retargeting. In *Computer Graphics Forum*, Vol. 19. Wiley Online Library, 11–19.

[37] Chong Mou, Xintao Wang, Liangbin Xie, Jian Zhang, Zhongang Qi, Ying Shan, and Xiaohu Qie. 2023. T2i-adapter: Learning adapters to dig out more controllable ability for text-to-image diffusion models. *arXiv preprint arXiv:2302.08453* (2023).

[38] Haomiao Ni, Changhao Shi, Kai Li, Sharon X Huang, and Martin Renqiang Min. 2023. Conditional Image-to-Video Generation with Latent Flow Diffusion Models. In *Proceedings of the IEEE/CVF Conference on Computer Vision and Pattern Recognition*. 18444–18455.

[39] Junting Pan, Chengyu Wang, Xu Jia, Jing Shao, Lu Sheng, Junjie Yan, and Xiaogang Wang. 2019. Video generation from single semantic label map. In *Proceedings of the IEEE/CVF Conference on Computer Vision and Pattern Recognition*. 3733–3742.

[40] Xingang Pan, Ayush Tewari, Thomas Leimkühler, Lingjie Liu, Abhimitra Meka, and Christian Theobalt. 2023. Drag your gan: Interactive point-based manipulation on the generative image manifold. In *ACM SIGGRAPH 2023 Conference Proceedings*. 1–11.

[41] Xue Bin Peng, Pieter Abbeel, Sergey Levine, and Michiel van de Panne. 2018. DeepMimic: Example-Guided Deep Reinforcement Learning of Physics-Based Character Skills. *ACM Trans. Graph. (SIGGRAPH)* (2018).

[42] Sigal Raab, Inbal Leibovitch, Guy Tevet, Moab Arar, Amit H. Bermano, and Daniel Cohen-Or. 2023. Single Motion Diffusion. arXiv:2302.05905 [cs.CV]

[43] Robin Rombach, Andreas Blattmann, Dominik Lorenz, Patrick Esser, and Björn Ommer. 2022. High-resolution image synthesis with latent diffusion models. In *IEEE Conf. Comput. Vis. Pattern Recog.* 10684–10695.

[44] Olaf Ronneberger, Philipp Fischer, and Thomas Brox. 2015. U-net: Convolutional networks for biomedical image segmentation. In *Medical Image Computing and Computer-Assisted Intervention–MICCAI 2015: 18th International Conference, Munich, Germany, October 5-9, 2015, Proceedings, Part III 18*. Springer, 234–241.

[45] Nataniel Ruiz, Yuanzhen Li, Varun Jampani, Yael Pritch, Michael Rubinstein, and Kfir Aberman. 2023. Dreambooth: Fine tuning text-to-image diffusion models for subject-driven generation. In *Proceedings of the IEEE/CVF Conference on Computer Vision and Pattern Recognition*. 22500–22510.

[46] Simo Ryu. 2023. Low-rank adaptation for fast text-to-image diffusion fine-tuning.

[47] Yujun Shi, Chuhui Xue, Jiachun Pan, Wenqing Zhang, Vincent YF Tan, and Song Bai. 2023. DragDiffusion: Harnessing Diffusion Models for Interactive Point-based Image Editing. *arXiv preprint arXiv:2306.14435* (2023).

[48] Aliaksandr Siarohin, Stéphane Lathuilière, Sergey Tulyakov, Elisa Ricci, and Nicu Sebe. 2019. First order motion model for image animation. In *Adv. Neural Inform.*

[49] Aliaksandr Siarohin, Stéphane Lathuilière, Sergey Tulyakov, Elisa Ricci, and Nicu Sebe. 2019. Animating Arbitrary Objects via Deep Motion Transfer. In *IEEE Conf. Comput. Vis. Pattern Recog.*

[50] Aliaksandr Siarohin, Oliver J Woodford, Jian Ren, Menglei Chai, and Sergey Tulyakov. 2021. Motion representations for articulated animation. In *IEEE Conf. Comput. Vis. Pattern Recog.* 13653–13662.

[51] Uriel Singer, Adam Polyak, Thomas Hayes, Xi Yin, Jie An, Songyang Zhang, Qiyuan Hu, Harry Yang, Oron Ashual, Oran Gafni, et al. 2022. Make-a-video: Text-to-video generation without text-video data. *arXiv preprint arXiv:2209.14792* (2022).

[52] Jiaming Song, Chenlin Meng, and Stefano Ermon. 2021. Denoising Diffusion Implicit Models. In *Int. Conf. Learn. Represent.*

[53] Seyoon Tak and Hyeong-Seok Ko. 2005. A physically-based motion retargeting filter. *ACM Transactions on Graphics (TOG)* 24, 1 (2005), 98–117.

[54] Luming Tang, Menglin Jia, Qianqian Wang, Cheng Perng Phoo, and Bharath Hariharan. 2023. Emergent Correspondence from Image Diffusion. *arXiv preprint arXiv:2306.03881* (2023).

[55] Jiale Tao, Biao Wang, Tiezheng Ge, Yuning Jiang, Wen Li, and Lixin Duan. 2022. Motion Transformer for Unsupervised Image Animation. In *Eur. Conf. Comput. Vis.* Springer, 702–719.

[56] Jiale Tao, Biao Wang, Borun Xu, Tiezheng Ge, Yuning Jiang, Wen Li, and Lixin Duan. 2022. Structure-aware motion transfer with deformable anchor model. In *IEEE Conf. Comput. Vis. Pattern Recog.* 3637–3646.

[57] Guy Tevet, Sigal Raab, Brian Gordon, Yonatan Shafir, Daniel Cohen-Or, and Amit H. Bermano. 2023. Human Motion Diffusion Model. In *Int. Conf. Learn. Represent.*

[58] Tan Wang, Linjie Li, Kevin Lin, Chung-Ching Lin, Zhengyuan Yang, Hanwang Zhang, Zicheng Liu, and Lijuan Wang. 2023. Disco: Disentangled control for referring human dance generation in real world. *arXiv e-prints* (2023), arXiv–2307.

[59] Yaohui Wang, Xinyuan Chen, Xin Ma, Shangchen Zhou, Ziqi Huang, Yi Wang, Ceyuan Yang, Yinan He, Jiashuo Yu, Peiqing Yang, et al. 2023. Lavie: High-quality video generation with cascaded latent diffusion models. *arXiv preprint arXiv:2309.15103* (2023).

[60] Yaohui Wang, Xin Ma, Xinyuan Chen, Antitza Dantcheva, Bo Dai, and Yu Qiao. 2023. LEO: Generative Latent Image Animator for Human Video Synthesis. *arXiv preprint arXiv:2305.03989* (2023).

[61] Jay Zhangjie Wu, Yixiao Ge, Xintao Wang, Stan Weixian Lei, Yuchao Gu, Yufei Shi, Wynne Hsu, Ying Shan, Xiaohu Qie, and Mike Zheng Shou. 2023. Tune-a-video: One-shot tuning of image diffusion models for text-to-video generation. In *IEEE Conf. Comput. Vis. Pattern Recog.* 7623–7633.

[62] Ruiqi Wu, Liangyu Chen, Tong Yang, Chunle Guo, Chongyi Li, and Xiangyu Zhang. 2023. Lamp: Learn a motion pattern for few-shot-based video generation. *arXiv preprint arXiv:2310.10769* (2023).

[63] Ling Yang, Zhilong Zhang, Yang Song, Shenda Hong, Runsheng Xu, Yue Zhao, Wentao Zhang, Bin Cui, and Ming-Hsuan Yang. 2023. Diffusion Models: A Comprehensive Survey of Methods and Applications. arXiv:2209.00796 [cs.LG]

[64] Jiangning Zhang, Chao Xu, Liang Liu, Mengmeng Wang, Xia Wu, Yong Liu, and Yunliang Jiang. 2020. Dtvnet: Dynamic time-lapse video generation via single still image. In *Computer Vision–ECCV 2020: 16th European Conference, Glasgow, UK, August 23–28, 2020, Proceedings, Part V 16.* Springer, 300–315.

[65] Jian Zhao and Hui Zhang. 2022. Thin-plate spline motion model for image animation. In *IEEE Conf. Comput. Vis. Pattern Recog.* 3657–3666.

[66] Daquan Zhou, Weimin Wang, Hanshu Yan, Weiwei Lv, Yizhe Zhu, and Jiashi Feng. 2022. Magicvideo: Efficient video generation with latent diffusion models. *arXiv preprint arXiv:2211.11018* (2022).

[67] Jun-Yan Zhu, Taesung Park, Phillip Isola, and Alexei A Efros. 2017. Unpaired image-to-image translation using cycle-consistent adversarial networks. In *Proceedings of the IEEE international conference on computer vision.* 2223–2232.

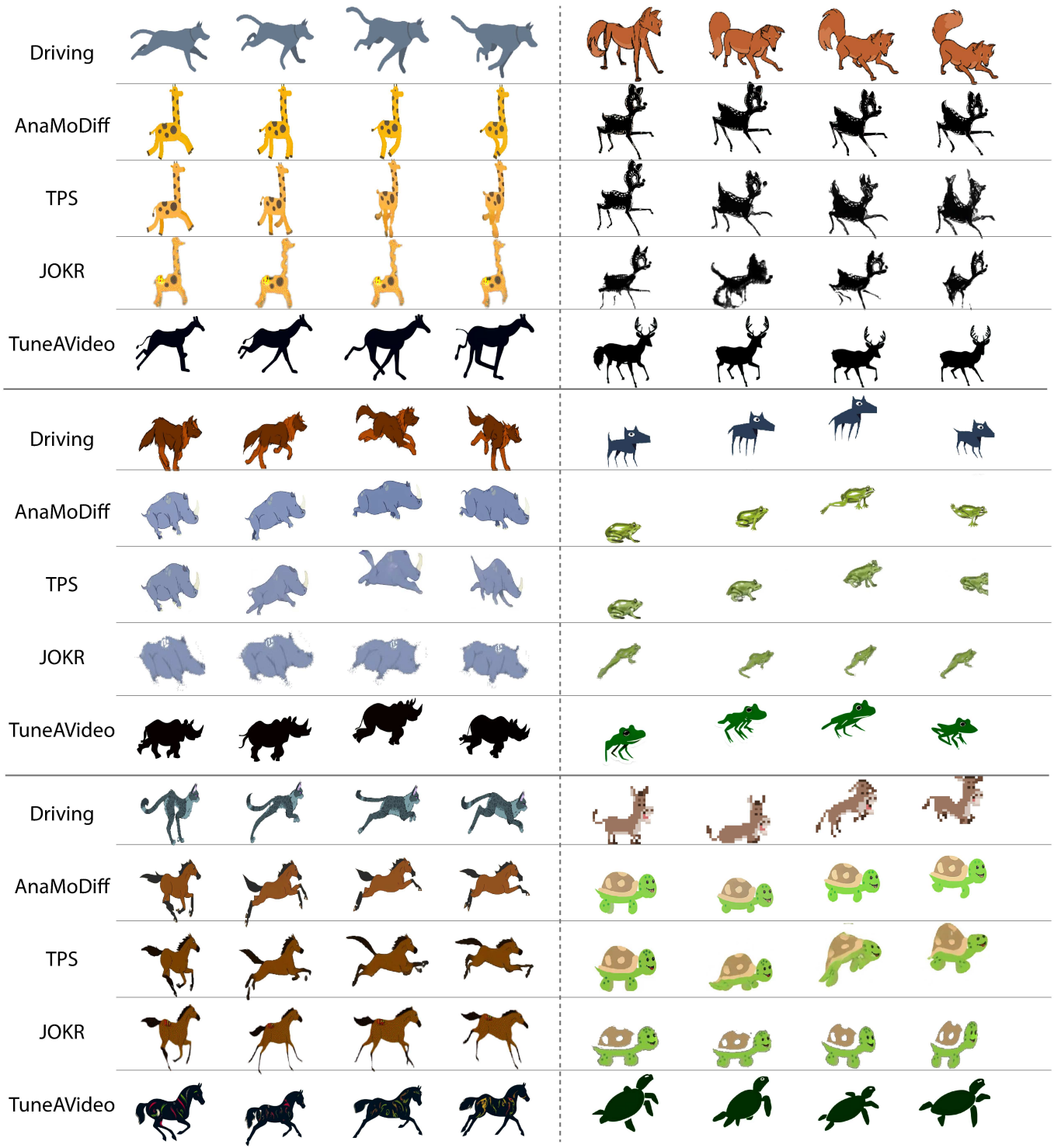

Fig. 9. More animal results compared with TPS, JOKR and TuneAVideo.
10

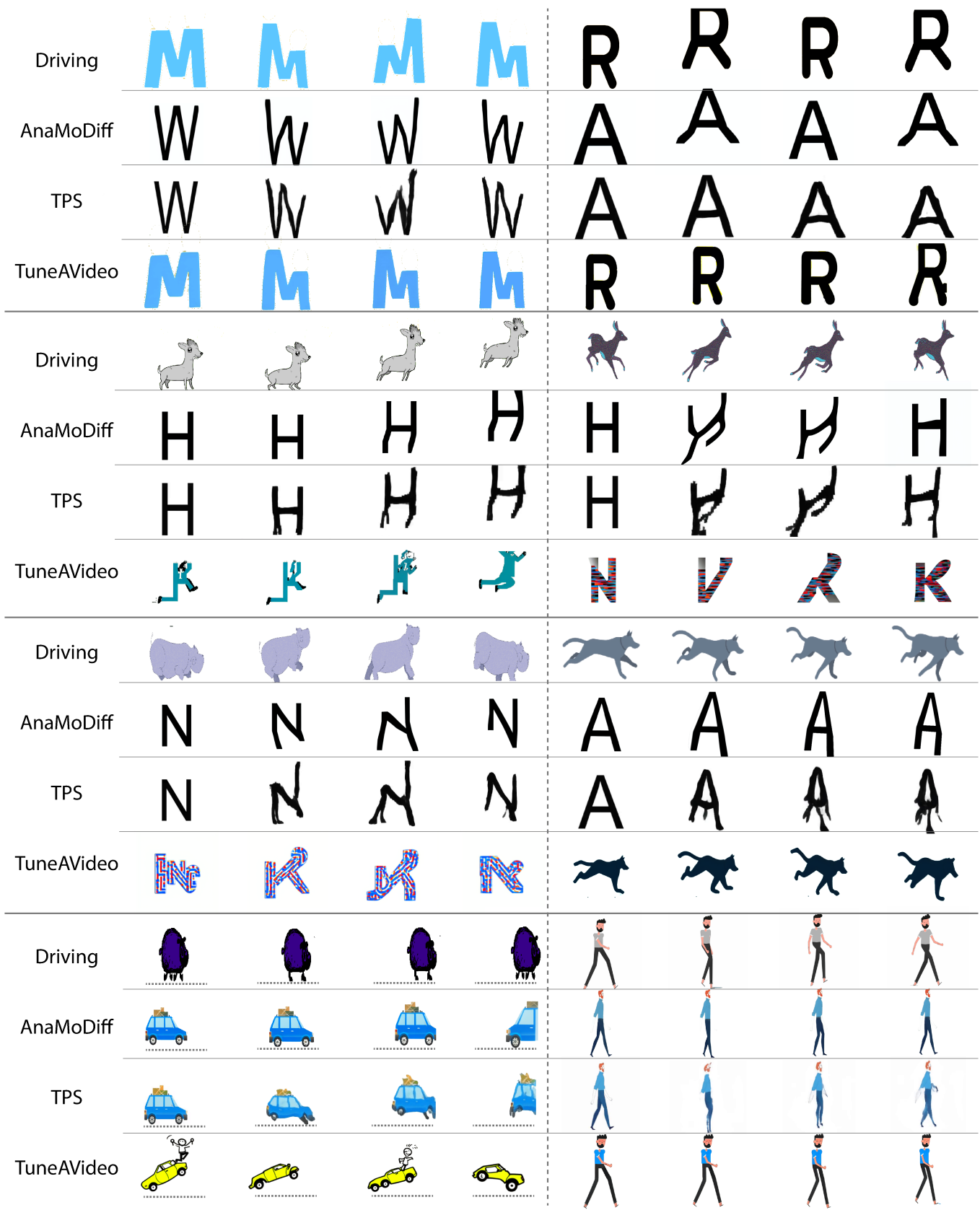

Fig. 10. We show some non-animal results. The top row shows results for transferring motion from letter animation to other letters. The next two rows show animal motion to letter results. In the last row, we show an animal-to-car and a human-to-human motion transfer.

11